%% file: emnlp2023.tex
\pdfoutput=1

\documentclass[11pt]{article}

\usepackage[]{EMNLP2023}

\usepackage{times}
\usepackage{latexsym}

\usepackage[T1]{fontenc}

\usepackage[utf8]{inputenc}

\usepackage{microtype}
\usepackage{xcolor}         
\usepackage{multirow}
\usepackage{subfigure}
\usepackage{xcolor}
\usepackage{bbding}
\usepackage{amsmath}
\usepackage{graphicx}
\usepackage[utf8]{inputenc} 
\usepackage[T1]{fontenc}    
\usepackage{hyperref}       
\usepackage{url}            
\usepackage{booktabs}       
\usepackage{amsfonts}       
\usepackage{nicefrac}       
\usepackage{microtype}
\usepackage{color}
\definecolor{dt}{gray}{0.5}  

\usepackage{colortbl}
\usepackage{subcaption}
\usepackage{pifont}

\usepackage{inconsolata}

\usepackage{enumitem}

\title{RWKV-CLIP: A Robust Vision-Language Representation Learner}

\author{Tiancheng Gu$^\text{\ding{170}}$\footnotemark[1], 
Kaicheng Yang$^{\text{\ding{171}}}$\footnotemark[1], 
Xiang An$^{\text{\ding{171}}}$, 
Ziyong Feng$^{\text{\ding{171}}}$ \\ 
\textbf{Dongnan Liu}$^{\text{\ding{170}}}$, 
\textbf{Weidong Cai}$^{\text{\ding{170}}}$\footnotemark[2],
\textbf{Jiankang Deng}$^{\text{\ding{105}}}$\footnotemark[2]\\
$^{\text{\ding{170}}}$University of Sydney $^{\text{\ding{171}}}$DeepGlint $^{\text{\ding{105}}}$Imperial College \\
\texttt\small{\small tigu8498@uni.sydney.edu.au, \small kaichengyang@deepglint.com}}

\begin{document}
\maketitle

\renewcommand{\thefootnote}{\fnsymbol{footnote}}
\footnotetext[1]{Equal contribution.}
\footnotetext[2]{Corresponding author.}

\begin{abstract}
Contrastive Language-Image Pre-training (CLIP) has significantly improved performance in various vision-language tasks by expanding the dataset with image-text pairs obtained from the web. This paper further explores CLIP from the perspectives of data and model architecture. To mitigate the impact of the noise data and enhance the quality of large-scale image-text data crawled from the internet, we introduce a diverse description generation framework that can leverage Large Language Models~(LLMs) to combine and refine information from web-based image-text pairs, synthetic captions, and detection tags. Additionally, we propose RWKV-CLIP, the first RWKV-driven vision-language representation learning model that combines the effective parallel training of transformers with the efficient inference of RNNs. Extensive experiments across different model scales and pre-training datasets demonstrate that RWKV-CLIP is a robust vision-language representation learner and it achieves state-of-the-art performance across multiple downstream tasks, including linear probing, zero-shot classification, and zero-shot image-text retrieval.  To facilitate future research, the code and pre-trained models are released at~\url{https://github.com/deepglint/RWKV-CLIP}.
\end{abstract}

\input{section/introduction}
\input{section/releted_work}
\input{section/method}
\input{section/experiment}

\input{section/ablation}

\input{section/conclusion}

\section{Limitations}
Our proposed framework for diverse description generation leverages the existing caption generation model and detection tags model, both of which can directly influence the quality of the final generated descriptions. Furthermore, due to limitations in computational resources, this study only executes experiments at tens of millions of scales of image-text pairs. Conducting experiments at a billion-scale necessitates substantial computational resources.

\bibliography{custom}
\bibliographystyle{acl_natbib}

\clearpage
\input{appendix}

\end{document}

%% file: section/introduction.tex
\section{Introduction}
The proliferation of mobile networks and social platforms has greatly accelerated the large-scale production of image-text pairs~\cite{yang2020cm, yu2020ch}. This unprecedented abundance of data has established the foundation for vision-language pre-training. Contrastive Language-Image Pre-training (CLIP) employs two distinct unimodal encoders for images and text, utilizing a contrastive loss, a highly effective mechanism for representation learning. Having been pre-trained on extensive image-text pairs collected from the internet, CLIP demonstrates strong transferability and has been widely applied across various domains~\cite{zhou2022zegclip,yao2023detclipv2}.

\begin{figure}[t]
\centering
{\includegraphics[width=0.23\textwidth]{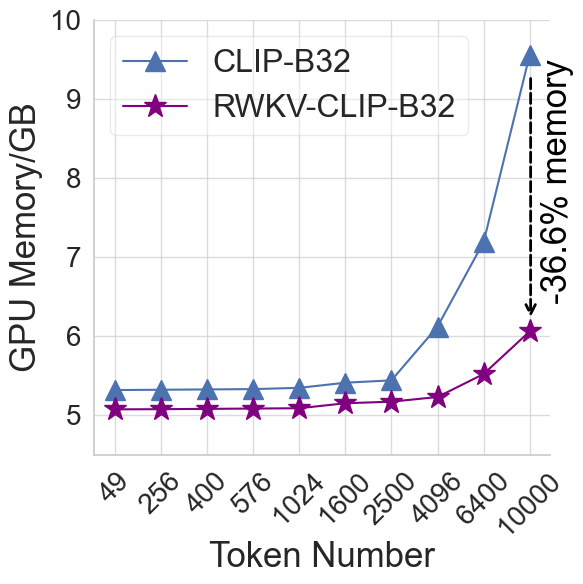}}
{\includegraphics[width=0.23\textwidth]{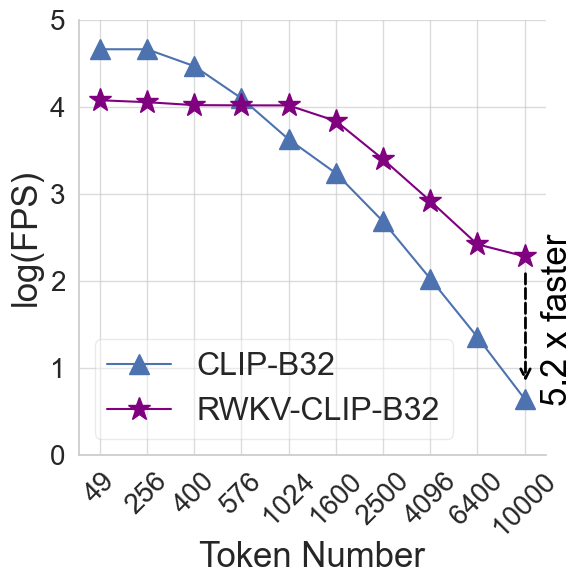}}
{\includegraphics[height=0.23\textwidth]{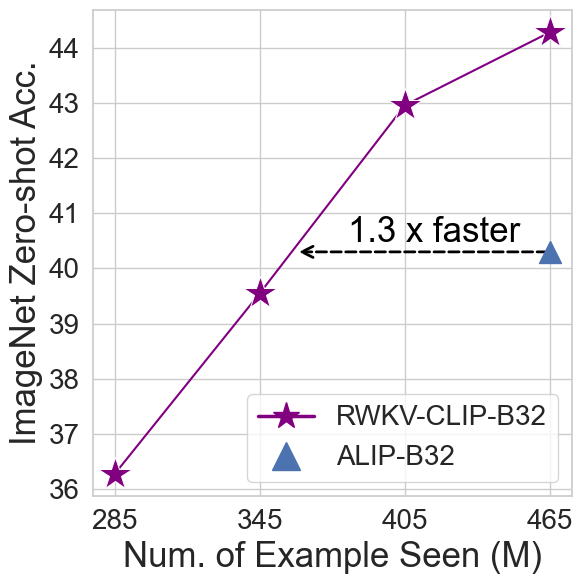}}
{\includegraphics[height=0.23\textwidth]{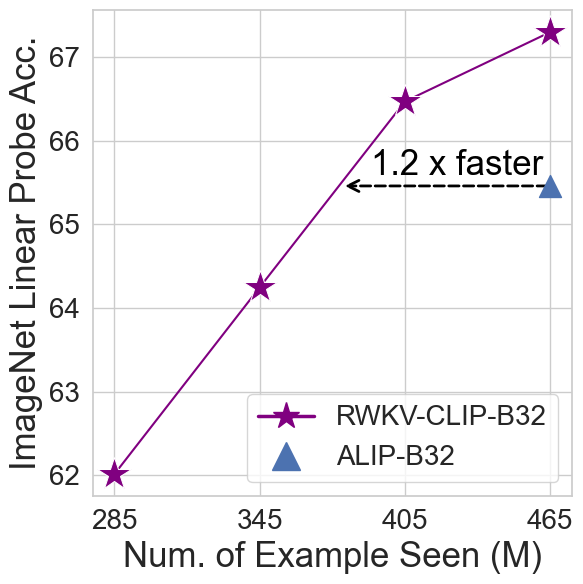}}
\caption{The proposed RWKV-CLIP combines the effective parallel training of transformers with the efficient inference of RNNs, achieving better efficiency and accuracy than the baseline methods (e.g., CLIP and ALIP).}
\vspace{-7mm}
\label{efficient}
\end{figure}

In recent years, many large-scale image-text datasets collected from the internet have been released. LAION400M~\cite{schuhmann2021laion} is created for research purposes and it contains 400 million image-text pairs curated using the CLIP model. Building on this, LAION5B~\cite{schuhmann2022laion}, which consists of 5.85 billion CLIP-filtered image-text pairs, successfully replicates and fine-tunes basic models such as CLIP. However, using the CLIP model to filter web-based image-text pairs still retains a considerable presence of noisy data. To improve data quality, DataComp~\cite{gadre2024datacomp} employs various strategies such as basic filtering, CLIP score filtering, and text\&image-based filtering. However, inherent characteristics of internet data, such as abstract text representations and semantic discrepancies between text and images, remain significant obstacles.

In recent years, the Transformer~\cite{vaswani2017attention} model has been extensively applied in large-scale representation learning, yielding significant performance improvements across multiple downstream tasks~\cite{acosta2022multimodal, kirillov2023segment, wang2023internimage}, including image classification~\cite{dosovitskiy2020image, Wang_2023_CVPR}, text generation~\cite{brown2020language}, and speech recognition~\cite{radford2023robust}. Despite these achievements, the quadratic computational complexity inherent in Transformer limits its capacity to effectively process high-resolution images and long sequences, posing a substantial challenge to its broader applicability across varied domains.

In this paper, we design a framework for generating diverse descriptions. Following ALIP~\cite{alip_Yang_2023_ICCV}, we first use the OFA~\cite{wang2022ofa} model to generate synthetic descriptions consistent with image content. However, constrained by the training data, OFA can only partially identify coarse-grained object categories. Therefore, we introduce an open-set image tagging model RAM++~\cite{huang2023open} to capture more detailed and precise semantic information from images. By leveraging LLMs, we synthesize and refine information from web-based texts, synthetic captions, and detection tags. Additionally, inspired by RWKV~\cite{peng2024eagle} and Vision-RWKV~\cite{duan2024vision}, we propose RWKV-CLIP, the first RWKV-driven vision-language representation learning model that combines the effective parallel training of Transformers with the efficient inference of RNNs. Extensive experiments across various model scales and pre-training datasets demonstrate that RWKV-CLIP is a robust and efficient vision-language representation learner. The main contributions of this paper are summarized as follows:
\begin{itemize}[noitemsep,topsep=0pt]
\item We introduce a diverse description generation framework, which can leverage LLMs to synthesize and refine information from web-based texts, synthetic captions, and detection tags to produce more accurate and semantically enriched descriptions.
\item We propose the RWKV-CLIP, the first RWKV-driven vision-language representation learning model, which combines the parallel training effectiveness of Transformers with the inference efficiency of RNNs.
\item We demonstrate the robustness and effectiveness of RWKV-CLIP as a vision-language representation learner through extensive experiments across various model scales and pre-training datasets.
\end{itemize}

%% file: section/releted_work.tex
\section{Related Work}
\begin{figure*}[t]
\centering
  \includegraphics[width=0.95\linewidth]{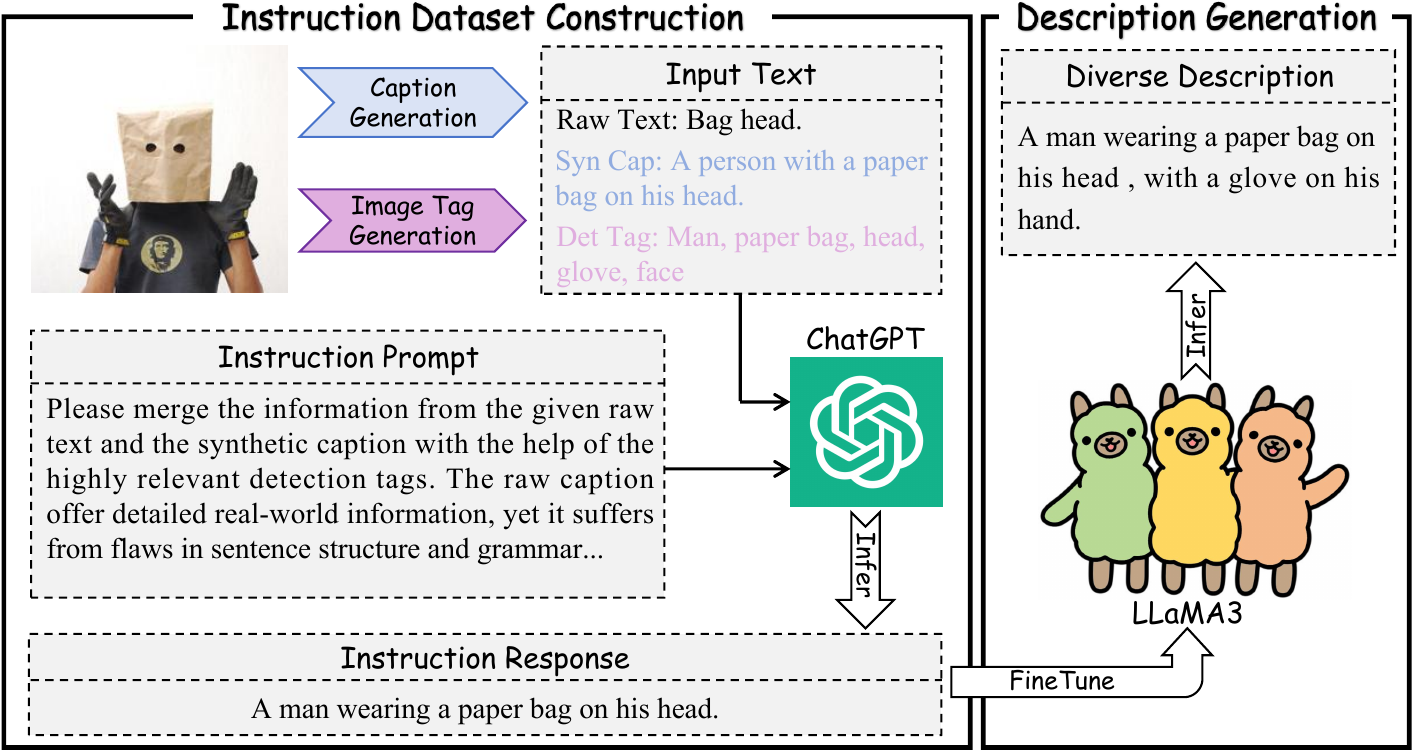}
  \caption{The architecture of our proposed diverse description generation framework.}
  \label{fig:rewrite}
\vspace{-5mm}
\end{figure*}

\subsection{Vision-Language Representation Learning}
As the milestone in vision-language representation learning, CLIP~\cite{radford2021learning} has garnered unparalleled interest due to its remarkable zero-shot recognition capability and outstanding transfer performance. Subsequently, a significant amount of enhancement works~\cite{yang2024clip,an2024multi,an2023unicom} based on CLIP have been proposed. SLIP~\cite{mu2022slip} combines self-supervised learning with CLIP pre-training to achieve significant performance improvements. DeCLIP~\cite{li2021supervision} employs multi-view supervision across modalities and nearest-neighbor supervision from similar pairs to enhance representation learning efficiency. FILIP~\cite{yao2021filip} refines contrastive loss to learn fine-grained representations for image patches and sentence words. UniCLIP~\cite{lee2022uniclip} boosts data efficiency by integrating contrastive loss across multiple domains into a single universal space. HiCLIP~\cite{geng2023hiclip} enhances cross-modal alignment by incorporating hierarchy-aware attention into both visual and language branches of CLIP. ALIP~\cite{alip_Yang_2023_ICCV} introduces a gating mechanism to reduce the influence of noisy pairs using synthetic data. Different from the above methods, this paper conducts further exploration of both the data and model architecture, proposing a diverse description generation framework and introducing RWKV-CLIP, the first RWKV-driven vision-language representation model.

\subsection{Text Agumentation}
With the success of LLMs in Natural Language Processing~(NLP), there is growing interest in leveraging LLMs to enhance text descriptions in large-scale image-text pairs. LaCLIP~\cite{fan2023improving} explores different strategies to generate rewrite examples and uses the in-context learning ability of LLMs to rewrite text within image-text datasets. However, the hallucination issue of LLMs and reliance on limited samples to guide the rewriting process can still introduce significant noise. To address this, CapsFusion~\cite{yu2023capsfusion} generates synthetic captions for each image and utilizes ChatGPT to merge raw texts and synthetic captions, creating a dataset with one million instructions for LLaMA fine-tuning. Despite this, caption generation models such as OFA~\cite{wang2022ofa} and BLIP~\cite{li2022blip} are limited by their training data and can only identify a restricted set of coarse-grained object categories. In this paper, we introduce the open-set image tagging model RAM++~\cite{huang2023open} to assign semantic detection tags to each image. Beneficial from detection tags, more semantic information can be introduced from images, which in turn further constrains LLMs and mitigates hallucinations.

\subsection{Receptance Weighted Key Value}
RWKV~\cite{peng2023rwkv} is first proposed in NLP, it addresses memory bottleneck and quadratic scaling in Transformers through efficient linear scaling while retaining expressive characteristics like parallelized training and robust scalability. Recently, Vision-RWKV~\cite{duan2024vision} successfully transferred the RWKV from NLP to vision tasks, outperforming ViT in image classification with faster processing and reduced memory consumption for high-resolution inputs. PointRWKV~\cite{he2024pointrwkv} demonstrates leading performance across various downstream tasks, surpassing Transformer- and Mamba-based counterparts in efficiency and computational complexity. Furthermore, Diffusion-RWKV~\cite{fei2024diffusion} adapts RWKV for diffusion models in image generation tasks, achieving competitive or superior performance compared to existing CNN or Transformer-based diffusion models. However, these methods have only validated RWKV in specific downstream tasks, and the potential of RWKVs to replace ViTs in vision-language representation learning remains unverified.

%% file: section/method.tex
\section{Method}

\begin{figure*}[t]
\centering
  \includegraphics[width=0.95\linewidth]{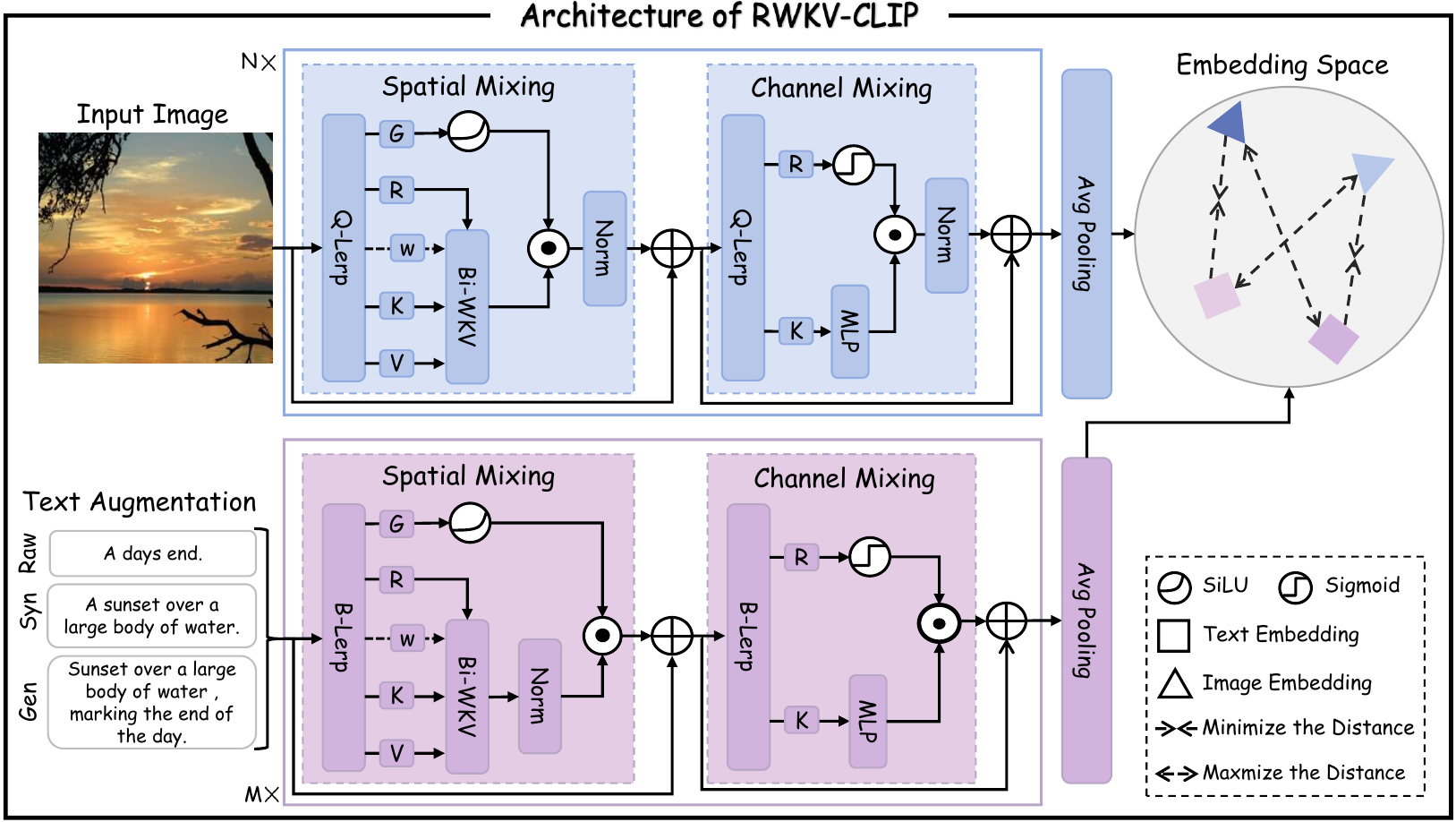}
  \caption{The architecture of RWKV-CLIP, which consists of M$\times$ and N$\times$ RWKV-driven blocks followed by an average pooling layer.}
  \label{fig:rwkv-clip}
\vspace{-5mm}
\end{figure*}

\label{section:method}
In this section, we first introduce a diverse description generation framework that leverages the capabilities of large language models to integrate information from web-based texts, synthetic captions, and detection tags. Subsequently, we provide a detailed exposition of RWKV-CLIP.

\subsection{Diverse Description Generation}
The architecture of our proposed diverse description generation framework is illustrated in Fig.~\ref{fig:rewrite}. To mitigate the effects of mismatched image-text pairs, following ALIP~\cite{alip_Yang_2023_ICCV}, we first adopt the OFA$_{base}$ model to generate a synthetic caption for each image. The synthetic captions exhibit a high degree of semantic alignment with the image, facilitating alignment across different modal feature spaces. However, constrained by the training data, OFA$_{base}$ can recognize a limited number of object categories and tends to produce captions with a simplistic sentence structure. To capture finer-grained semantic information within images, we incorporate the open-set image tagging models RAM++~\cite{huang2023open} to extract object detection tags for each image. 

To assess the viability of our approach, following CapsFusion~\cite{yu2023capsfusion}, we initially leverage ChatGPT to combine information from raw texts, synthetic captions, and detection tags. However, the time and computational effort involved is prohibitive. Therefore, we construct an instruction dataset based on ChatGPT interactions and fine-tuned the open-source LLaMA3 with this dataset. After that, we leverage the fine-tuned LLaMA3 model for large-scale inference. Specifically, we select 70K image-text pairs from YFCC15M that have more than 10 detection tags. Then, we input the raw texts, synthetic captions, and detection tags of these data into ChatGPT to get instruction responses. The details of the instruction prompt are provided in the supplementary material.

After obtaining the instruction dataset, we utilize the LLaMA Factory~\cite{zheng2024llamafactory} to finetune the LLaMA3-8B and leverage vLLM~\cite{kwon2023efficient} to accelerate large-scale inference.

\subsection{RWKV-CLIP}
In this section, we propose RWKV-CLIP, a robust and efficient RWKV-driven vision-language representation learner. Inspired by CLIP~\cite{radford2021learning} and Vision-RWKV~\cite{duan2024vision}, RWKV-CLIP adopts a dual-tower architecture with a block-stacked encoder design like the Transformer~\cite{vaswani2017attention}, where each block consists of a spatial mixing and a channel mixing module. The overview architecture of our proposed RWKV-CLIP is shown in Fig.~\ref{fig:rwkv-clip}. 

\noindent{\bf Input Augmentation.} Based on our proposed diverse description generation framework, we can obtain three types of text: raw text $T_{r}$, synthetic caption $T_{s}$, and generated description $T_{g}$. To improve the robustness of the model, we randomly select a text from $[T_{r}, T_{s}, T_{g}]$ as the augmentation for text inputs:
\begin{equation}
    \small 
    \mathrm{aug}(T) = \mathrm{Sample}([T_{r}, T_{s}, T_{g}]).
\end{equation}
Meanwhile, the input image $I \in \mathbb{R}^{H\times W \times 3}$ is transformed into $HW/p^2$ patches, where $p$ is the patch size.

\noindent{\bf Spatial Mixing.} The input text $\text{aug}(T)$ and image $I$ are passed through the spatial mixing module, which acts as an attention mechanism and performs global attention computation of linear complexity. Specifically, the input data is shifted and entered into four parallel linear layers to obtain multi-head vectors $G_{x}^{s}, R_{x}^{s}, K_{x}^{s}, V_{x}^{s}$:
\begin{equation}
    \small 
    \psi_{x}^{s} = \mathrm{Lerp}_{\psi}(x) \cdot w_{\psi}^{s}, \quad \psi \in \{G, R, K, V\},
\end{equation}
where $\rm Lerp$ is the linear interpolation~\cite{peng2024eagle}. In this paper, we adopt $\text{Q-Lerp}$ and $\text{B-Lerp}$ for image and text encoders respectively. The $\text{Q-Lerp}$ can be formulated as:
\begin{equation}
\small
\begin{aligned}
    &\text{Q-Lerp}_{\Psi}(I) = I + (1 - \eta_{\Psi}) \cdot I^{\star}, \\
    &I^{\star} = \text{Concat}(I_{1}, I_{2}, I_{3}, I_{4}).
\end{aligned}
\end{equation}

The $\text{B-Lerp}$ can be presented as:
\begin{equation}
\small
\begin{aligned}
    &\text{B-Lerp}_{\Psi}(T) = T + (1 - \eta_{\Psi}) \cdot T^{\star}, \\
    &T^{\star} = \text{Concat}(T_{1}, T_{2}),
\end{aligned}
\end{equation}

where $\Psi \in \{G, R, K, V, w\}$, $\eta_{\Psi}$ denotes learnable vectors, $I^{\star}$ is the quad-directional shift vector in the image, \emph{i.e.}, $I_{1} = x[h-1,w,0:C/4], I_{2} = x[h+1,w,C/4:C/2], I_{3} = x[h,w-1,C/2:3C/4], I_{4} = x[h,w+1,3C/4:C]$, $T^{\star}$ is the bi-directional shift in the text \emph{i.e.}, $T_{1} = [w-1, 0:C/2]$, $T_{2} = [w+1, C/2:C]$, where $h, w, C$ present the number of height, width, and channel. These shift functions enhance feature interaction at the channel level, enabling a focus on neighboring tokens. Specifically, the bi-directional shift ensures forward and backward interaction of text tokens without increasing additional FLOPs. To avoid a fixed learned vector, a new time-varying decay $w_{x}$ is calculated as follows:
\begin{equation}
\small
\begin{aligned}
    &\phi(x) = \lambda + \tanh(x \cdot M_{i}) \cdot M_{j}, \\
    &\hat{w}_{x}^{s} = x + (1 - \phi(\text{Lerp}_{w}(x))) \cdot x^{\star}, \\
    &\tilde{w}_{x}^{s} = \phi(\hat{w}_{x}^{s}), w_{x}^{s} = \exp\left(-\exp(\tilde{w}_{x}^{s})\right),
\end{aligned}
\end{equation}
where $x \in \{I,T\}$, $\lambda$ is a learnable vector, $M_{i}, M_{j}$ are learnable weight matrices. The function $\phi$ is used to obtain learned vectors by inexpensively augmenting inputs with additional offsets. $\hat{w}_{x}^{s}$ and $\Tilde{w}_{x}^{s}$ are middle values of $w_{x}^{s}$ during the calculation process. This process allows each channel of $w_{x}$ to vary based on a mix of the current and prior tokens $x^{\star}$.

Subsequently, $w_{x}^{s}, R_{x}^{s}, K_{x}^{s}, V_{x}^{s}$ are used to compute the global attention result $wkv_{t}$ via a linear complexity bidirectional attention mechanism. This result is then multiplied by $\sigma(G_{x}^{s})$, functioning as a gate mechanism to control the output $O_{x}^{s}$:
\begin{equation}
\small
\begin{aligned}
    &wkv_{t} = \text{Bi-WKV}_{t}(w_{x}^{s}, R_{x}^{s}, K_{x}^{s}, V_{x}^{s}), \\
    &O_{x}^{s} = \text{Concat}\left(\sigma(G_{x}^{s}) \odot \text{LN}(wkv_{t})\right) \cdot w_{o}^{s},
\end{aligned}
\end{equation}
where $\sigma(\cdot)$ denotes the SiLU function~\cite{elfwing2018sigmoid}, and $\odot$ means element-wise multiplication, $\rm LN$ is the layer norm and the \text{Bi-WKV}~\cite{duan2024vision, peng2024eagle} can be formulated as:
\begin{equation}
\small
\begin{aligned}
    \text{Bi-WKV}_{t} &= R_{s,t} \cdot \text{diag}(u) \cdot K_{s,t}^{\mathsf{T}} \cdot V_{s,t} \\
    &\quad + \sum_{i=0}^{t-1} \text{diag}(\epsilon_{i,j}) \cdot K_{s,i}^{\mathsf{T}} \cdot V_{s,i} \\
    &\quad + \sum_{i=t+1}^{T-1} \text{diag}(\epsilon_{i,j}) \cdot K_{s,i}^{\mathsf{T}} \cdot V_{s,i} ),
\end{aligned}
\end{equation}
where $u$ is a per-channel learned boost and $\epsilon_{i,j} = \odot^{i-1}_{j=1} w_{j}$ is a dynamic decay.

\noindent{\bf Channel Mixing.} The spatial mixing module is followed by the channel-mixing module. Similarly, the $R_{x}^{c}, K_{x}^{c}$ are obtained by Lerp:
\begin{equation}
\small
    \psi_{x}^{c} = \text{Lerp}_{\psi}(x) \cdot w_{\psi}^{c}, \quad \psi \in \{R, K\}.
\end{equation}
After that, a linear projection and a gate mechanism are performed respectively and the final output $O_{x}^{c}$ is formulated as:
\begin{equation}
\small
    O_{x}^{c} = (\sigma(R_{x}^{c}) \odot \rho(K_{x}^{c})) \cdot w_{o}^{c},
\end{equation}
where $\rho$ is the squaredReLU~\cite{agarap2018deep}. After passing through the stack RWKV-based image and text encoders $E_{I}$ and $E_{T}$, we can get the image embeddings $\hat{I} = E_{I}(I)$ and text embeddings $\hat{T} = E_{T}(aug(T))$, the loss function $L$ is defined as:
{
\begin{equation}
\small
\begin{split}
L = -\sum_{i=1}^N \left[\log\frac{e^{\hat{I}_{i}^\top \hat{T}_{i}/\tau}}{\sum_{j}e^{\hat{I}_{i}^\top \hat{T}_{j}/\tau}}+\log\frac{e^{\hat{I}_{i}^\top \hat{T}_{i}/\tau}}{\sum_{j}e^{\hat{I}_{j}^\top \hat{T}_{i}/\tau}}\right].
\end{split}
\end{equation}
}

%% file: section/experiment.tex
\section{Experiments}
\input{table/linear_prob}
\input{table/zero_shot_retrieval}

\subsection{Experimental Settings}
\noindent{\bf Pre-training Datasets.} We train our model on the YFCC15M dataset, which is a subset of YFCC100M~\cite{thomee2016yfcc100m} filtered by DeCLIP~\cite{li2021supervision}. To further verify the effectiveness and generalizability of RWKV-CLIP, following ALIP~\cite{alip_Yang_2023_ICCV}, we randomly select subsets of 10M and 30M from the LAION400M~\cite{schuhmann2021laion}. We then conduct a series of experiments with different model scales and pre-training datasets.

\noindent{\bf Implementation Details.} Consistent with ALIP~\cite{alip_Yang_2023_ICCV}, we employ OFA$_{base}$ to generate synthetic captions. The instruction dataset is constructed using ChatGPT-35-turbo, and we fine-tune LLaMA3-8B to enhance the generation of diverse descriptions. We employ AdamW~\cite{loshchilov2017decoupled} as the optimizer, initialized with a learning rate of $1e-3$ and a weight decay of $0.2$. The parameters $\beta{1}$ and $\beta_{2}$ are set to $0.9$ and $0.98$, respectively. The input image size is $224 \times 224$, and the input text sequence length is truncated or padded to $77$. The temperature parameter $\tau$ is initialized to $0.07$. We train RWKV-CLIP for 32 epochs with a batch size of $4096$ on $8$ NVIDIA A100(80G) GPUs. We meticulously regulate the parameters and FLOPs of RWKV-CLIP to ensure the fairness of the experimental comparison. Please refer to the supplementary material for more detailed parameters, FLOPs, and settings of RWKV-CLIP.

\subsection{Experimental Results}
\noindent{\bf Linear Probe.} Building upon previous works~\cite{alip_Yang_2023_ICCV,li2021supervision,geng2023hiclip}, we use RWKV-CLIP as a feature extractor and train only a logistic regression classifier. Tab.~\ref{tab:linear} details the linear probe performance across 10 downstream datasets, as referenced in ALIP~\cite{alip_Yang_2023_ICCV}. RWKV-CLIP achieves a significant performance improvement ranging from 1.9\% $\sim$ 11.1\% over the baseline models, outperforming ALIP in 8 of the 10 datasets. The observed performance improvements are primarily due to two main factors: (1) Our proposed description generation framework effectively synthesizes and refines information from web-based texts, synthetic captions, and detection tags, producing more accurate and semanti-
cally enriched descriptions. (2) RWKV-CLIP exhibits superior representation learning capabilities compared to ViT-based models.

\noindent{\bf Zero-shot Image-text Retrieval.} In Tab.~\ref{tab:retrieval}, we compare our method with state-of-the-art approaches in zero-shot image-text retrieval on Flickr30k and MSCOCO. RWKV-CLIP achieves new state-of-the-art results on all evaluation metrics. Specifically, RWKV-CLIP achieves 76.0\%/57.6\%  I2T/T2I retrieval Recall@1 on Flickr30K, surpassing ALIP by 5.5\%/8.7\%. Similarly, significant improvements of 3.5\%/4.7\% in I2T/T2I retrieval Recall@1 are observed for RWKV-CLIP on MSCOCO. This exceptional image-text retrieval capability indicates that the representations learned by RWKV-CLIP are robust and exhibit enhanced cross-modal alignment.

\input{table/zero_shot_classification}

\noindent{\bf Zero-shot Classification.} We present the zero-shot classification performance across 11 datasets. To ensure fair comparisons, we use the same prompt templates and class names as established in ALIP~\cite{alip_Yang_2023_ICCV} and SLIP~\cite{mu2022slip}. As shown in Tab.~\ref{tab:transfer}, RWKV-CLIP achieves an average performance improvement of 2.6\% $\sim$ 14.4\% over baseline models. Notably, our model outperforms ALIP in 10 out of the 11 datasets, with significant enhancements on instance discrimination datasets such as Food101, and ImageNet. This improvement is mainly due to the diverse descriptions generated by our framework, providing more fine-grained semantic information.

\noindent{\bf Zero-Shot Robustness Evaluation.} In Tab.~\ref{tab:robustness}, we present a robustness evaluation comparing ALIP and RWKV-CLIP. Our results show that RWKV-CLIP consistently outperforms ALIP in terms of robustness across all datasets with an average improvement of 2.0\%. These experimental results establish the RWKV-driven model as a robust representation learner.

\begin{table}[h!]
\centering
\resizebox{\linewidth}{!}{
    \begin{tabular}{lccccc}
        \toprule

        Method  & IN-V2 & IN-A & IN-R  & IN-Sketch & Average\\
        \midrule
        ALIP-ViT-B/32   & 34.1 & 16.1 & 35.2 & 12.1 & 24.4  \\
        RWKV-CLIP-B/32  & \bf37.5 & \bf16.7 & \bf37.0 & \bf14.5 & \bf26.4 \\
	\bottomrule
    \end{tabular}
    }
\caption{Zero-shot robustness comparison of ALIP and RWKV-CLIP pretrained on YFCC15M.}
\label{tab:robustness}
\vspace{-5mm}
\end{table}

%% file: table/linear_prob.tex
\begin{table*}[t]
\centering
\resizebox{\linewidth}{!}{
    \begin{tabular}{lcccccccccccccc}
        \toprule
        Method & \shortstack{Pre-train \\ dataset} &  \rotatebox[origin=lb]{90}{\smash{CIFAR10}} & \rotatebox[origin=lb]{90}{\smash{CIFAR100}} &  \rotatebox[origin=lb]{90}{\smash{Food101}} & \rotatebox[origin=lb]{90}{\smash{Pets}} &  \rotatebox[origin=lb]{90}{\smash{Flowers}} & 
        \rotatebox[origin=lb]{90}{\smash{SUN397}} &
        \rotatebox[origin=lb]{90}{\smash{Cars}} & 
        \rotatebox[origin=lb]{90}{\smash{DTD}} & 
        \rotatebox[origin=lb]{90}{\smash{Caltech101}} & 
        \rotatebox[origin=lb]{90}{\smash{Aircraft}} & 
        \rotatebox[origin=lb]{90}{\smash{Average}}  \\
        \midrule

        CLIP-ViT-B/32\cite{radford2021learning} & YFCC15M & 86.5 & 64.7 & 69.2 & 64.6 & 90.6 & 66.0 & 24.9 & 61.3 & 79.1 & 23.1 & 63.0 \\
        DeCLIP-ViT-B/32~\cite{li2021supervision} & YFCC15M & 89.2 & 69.0 & 75.4 & 72.2 & 94.4 & 71.6 & 31.0 & 68.8 & 87.9 & 27.6 & 68.7 \\
        HiCLIP-ViT-B/32~\cite{geng2023hiclip}  & YFCC15M & 89.5 & 71.1 & 73.5 & 70.6 & 91.9 & 68.8 & 30.8 & 63.9 & 84.8 & 27.4 & 67.2 \\
        ALIP-ViT-B/32~\cite{alip_Yang_2023_ICCV} & YFCC15M & 94.3 & 77.8 & 75.8 & 76.0 & \textbf{95.1} & \textbf{73.3} & 33.6 & 71.7 & 88.5 & 36.1 & 72.2  \\
        \midrule
        RWKV-CLIP-B/32 & YFCC15M & \textbf{95.3} & \textbf{81.8} & \textbf{76.4} & \textbf{77.1} & 92.4 & 73.1 & \textbf{37.7} & \textbf{73.2} & \textbf{90.6} & \textbf{43.5} & \textbf{74.1}  \\
        \bottomrule
  \end{tabular}
  } 
\caption{Linear probe performance on 10 downstream datasets. RWKV-CLIP achieves an average performance improvement of 1.9\%$\sim$11.1\%.}
\label{tab:linear}
\vspace{-3mm}
\end{table*}

%% file: table/zero_shot_retrieval.tex
\begin{table*}[t]
\centering
\resizebox{\linewidth}{!}{
    \centering
    \begin{tabular}{lcccccccccccc}
    \toprule
    & \multicolumn{6}{c}{Text retrieval} & \multicolumn{6}{c}{Image retrieval} \\
    & \multicolumn{3}{c}{Flickr30k} & \multicolumn{3}{c}{MSCOCO} & \multicolumn{3}{c}{Flickr30k} & \multicolumn{3}{c}{MSCOCO} \\
    Method &  R@1 & R@5 & R@10 & R@1 & R@5 & R@10 & R@1 & R@5 & R@10 & R@1 & R@5 & R@10 \\
    \midrule
    CLIP-ViT-B/32\cite{radford2021learning} & 34.9 & 63.9 & 75.9 & 20.8 & 43.9 & 55.7 & 23.4 & 47.2 & 58.9 & 13.0 & 31.7 & 42.7 \\
    SLIP-ViT-B/32~\cite{mu2022slip} & 47.8 & 76.5 & 85.9 & 27.7 & 52.6 & 63.9 & 32.3 & 58.7 & 68.8 & 18.2 & 39.2 & 51.0 \\
    DeCLIP-ViT-B/32~\cite{li2021supervision} & 51.4 & 80.2 & 88.9 & 28.3 & 53.2 & 64.5 & 34.3 & 60.3 & 70.7 & 18.4 & 39.6 & 51.4 \\
    UniCLIP-ViT-B/32~\cite{lee2022uniclip} & 52.3 & 81.6 & 89.0 & 32.0 & 57.7 & 69.2 & 34.8 & 62.0 & 72.0 & 20.2 & 43.2 & 54.4 \\
    HiCLIP-ViT-B/32~\cite{geng2023hiclip} & - & - & - & 34.2 & 60.3 & 70.9 & - & - & - & 20.6 & 43.8 & 55.3 \\
    ALIP-ViT-B/32~\cite{alip_Yang_2023_ICCV} & 70.5 & 91.9 & 95.7 & 46.8 & 72.4 & 81.8 & 48.9 & 75.1 & 82.9 & 29.3 & 54.4 & 65.4 \\
    \midrule
    RWKV-CLIP-B/32 & \textbf{76.0} & \textbf{94.7} & \textbf{97.6} & \textbf{50.3} & \textbf{76.2} & \textbf{85.2} & \textbf{57.6} & \textbf{82.3} & \textbf{88.7} & \textbf{34.0} & \textbf{60.9} & \textbf{71.7} \\
    \bottomrule
    \end{tabular}
    }
\caption{Zero-shot image-text retrieval performance on the test splits of Flickr30k and MSCOCO. RWKV-CLIP achieves a significant improvement on all metrics.}
\label{tab:retrieval}
\vspace{-5mm}
\end{table*}

%% file: table/zero_shot_classification.tex
\begin{table*}[h!]
\centering
\resizebox{\linewidth}{!}{
    \begin{tabular}{lccccccccccccccc}
        \toprule
        
        Method & \shortstack{Pre-train \\ dataset} &  \rotatebox[origin=lb]{90}{\smash{CIFAR10}} & \rotatebox[origin=lb]{90}{\smash{CIFAR100}} &  \rotatebox[origin=lb]{90}{\smash{Food101}} & \rotatebox[origin=lb]{90}{\smash{Pets}} &  \rotatebox[origin=lb]{90}{\smash{Flowers}} & 
        \rotatebox[origin=lb]{90}{\smash{SUN397}} &
        \rotatebox[origin=lb]{90}{\smash{Cars}} & 
        \rotatebox[origin=lb]{90}{\smash{DTD}} & 
        \rotatebox[origin=lb]{90}{\smash{Caltech101}} & 
        \rotatebox[origin=lb]{90}{\smash{Aircraft}} & 
        \rotatebox[origin=lb]{90}{\smash{ImageNet}}  &
        \rotatebox[origin=lb]{90}{\smash{Average}}  \\
        \midrule

        CLIP-ViT-B/32\cite{radford2021learning} & YFCC15M & 63.7 & 33.2 & 34.6 & 20.1 & 50.1 & 35.7 & 2.6 & 15.5 & 59.9 & 1.2 & 32.8 & 31.8 \\

        SLIP-ViT-B/32~\cite{mu2022slip} & YFCC15M & 50.7 & 25.5 & 33.3 & 23.5 & 49.0 & 34.7 & 2.8 & 14.4 & 59.9 & 1.7 & 34.3 & 30.0 \\
        
        FILIP-ViT-B/32~\cite{yao2021filip} & YFCC15M & 65.5 & 33.5 & 43.1 & 24.1 & 52.7 & 50.7 & 3.3 & 24.3 & 68.8 & 3.2 & 39.5 & 37.2 \\
        
        DeCLIP-ViT-B/32~\cite{li2021supervision} & YFCC15M & 66.7 & 38.7 & 52.5 & 33.8 & 60.8 & 50.3 & 3.8 & \textbf{27.7} & 74.7 & 2.1 & 43.2 & 41.3 \\
        
        HiCLIP-ViT-B/32~\cite{geng2023hiclip} & YFCC15M & 74.1 & 46.0 & \textbf{51.2} & \textbf{37.8} & \textbf{60.9} & 50.6 & \textbf{4.5} & 23.1 & 67.4 & 3.6 & 40.5 & 41.8 \\
        ALIP-ViT-B/32~\cite{alip_Yang_2023_ICCV} & YFCC15M & \textbf{83.8} & 51.9 & 45.4 & 30.7 & 54.8 & 47.8 & 3.4 & 23.2 & 74.1 & 2.7 & 40.3 & 41.7  \\
        \midrule
        RWKV-CLIP-B/32 & YFCC15M & 79.8 & \textbf{55.1} & 50.6 & 37.6 & 57.1 & \textbf{54.0} & 4.1 & 24.6 & \textbf{77.1} & \textbf{4.0} & \textbf{44.3} & \textbf{44.4}  \\
        
        \bottomrule
  \end{tabular}
    } 
\caption{Zero-shot classification performance on 11 downstream datasets. RWKV-CLIP achieves an average performance improvement of 2.6\%$\sim$12.6\%.}
\label{tab:transfer}
\vspace{-3mm}
\end{table*}

%% file: section/ablation.tex
\subsection{Ablation Study}

\begin{figure*}[t]
\centering
    {
    \includegraphics[width=0.32\textwidth]{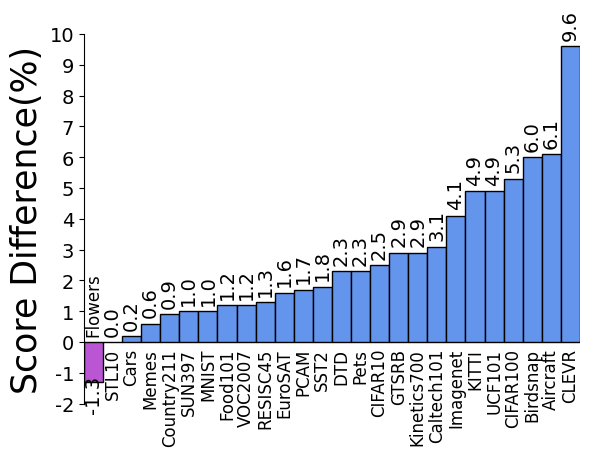}}
    {
    \includegraphics[width=0.32\textwidth]{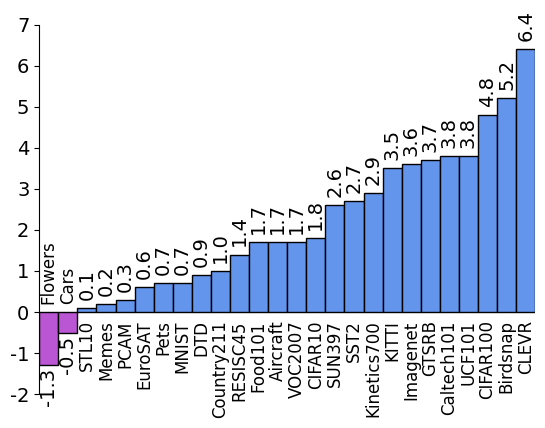}}
    {
    \includegraphics[width=0.32\textwidth]{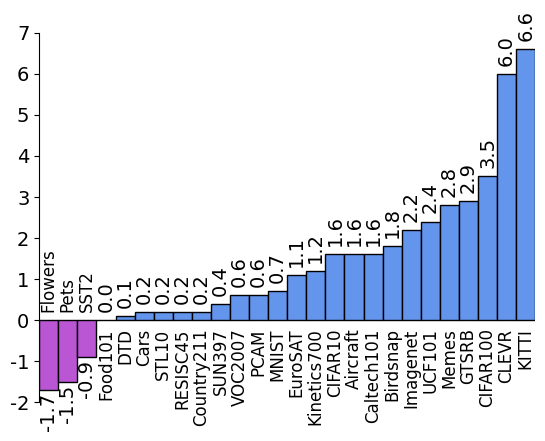}}
    \vspace{-3mm}
    \caption{Linear probe performance comparison between RWKV-CLIP and ALIP on 26 downstream datasets. The comparisons include RWKV-CLIP-B/32 vs. ALIP-ViT-B/32 on LAION10M, RWKV-CLIP-B/16 vs. ALIP-ViT-B/16 on LAION10M, and RWKV-CLIP-B/32 vs. ALIP-ViT-B/32 on LAION30M, presented from left to right.}
\label{fig:laion}
\vspace{-3mm}
\end{figure*}

\noindent{\bf Effectiveness of Model and Data Scaling.} To evaluate the effectiveness of RWKV-CLIP on model and data scaling, we conduct experiments on randomly selected subsets of 10M and 30M from LAION400M. For a more comprehensive comparison, we report the linear probe performance on 26 downstream datasets. As shown in Fig.~\ref{fig:laion}, RWKV-CLIP significantly improves performance across different model scales and pre-training datasets. These results demonstrate the robustness and extensibility of RWKV-CLIP. Detailed experimental results can be found in the supplementary material.

\begin{figure}[t!]
\centering
  \includegraphics[width=\linewidth]{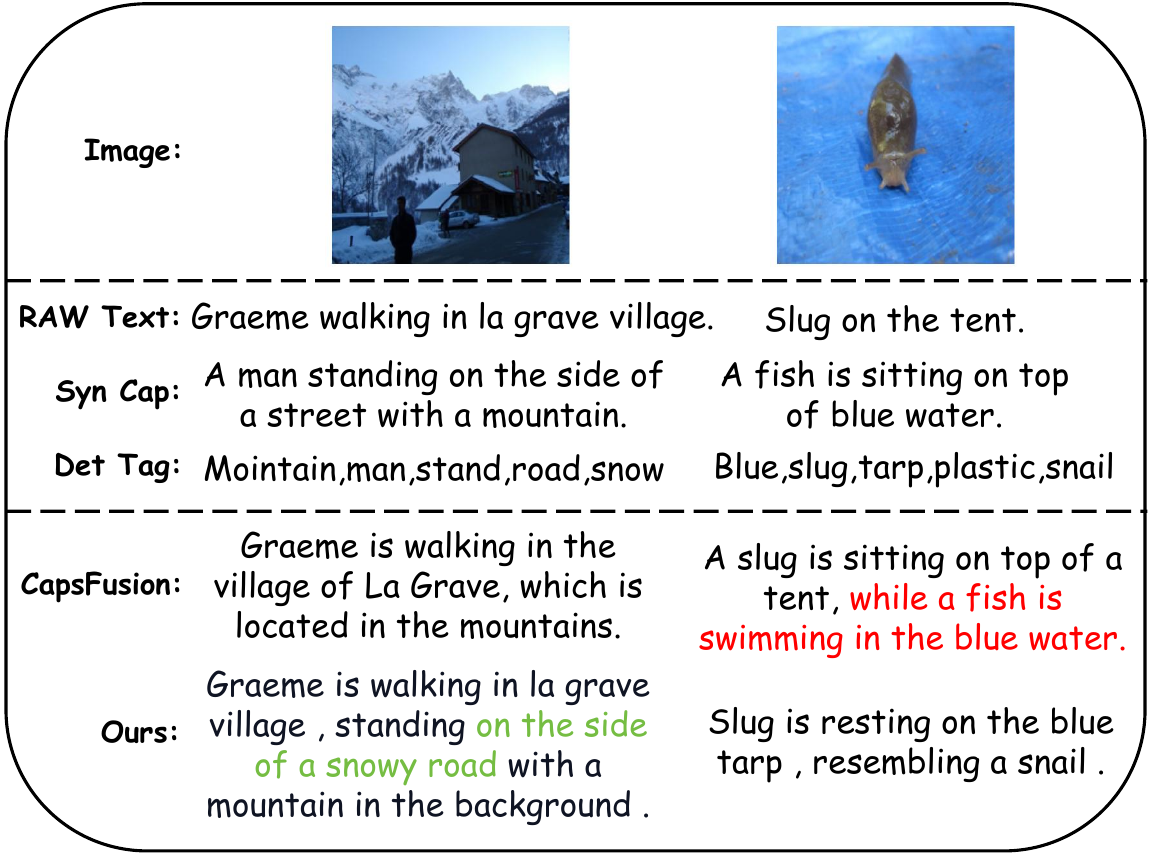}
  \caption{Comparison of our proposed diverse description generation framework vs. CapsFusion. Hallucinations are highlighted in \textcolor[HTML]{FF1E1E}{red}, and additional semantic information is highlighted in \textcolor[HTML]{6FBF50}{green}.}
\vspace{-3mm}
\label{fig:capsfusion_sample}
\end{figure}

\noindent{\bf Comparision Analysis with CapsFusion.} To further demonstrate the performance differences between our proposed diverse description generation framework and CapsFusion, we used CapsFusion-LLaMA to rewrite the YFCC15M dataset based on raw texts and synthetic captions. We then trained RWKV-CLIP using texts generated by our framework and CapsFusion. As shown in Tab.~\ref{tab:capsfusion_comparision}, our framework achieves a 0.9\% and 2.1\% improvement in the average linear probe and zero-shot classification performance, respectively. This improvement is primarily due to the detection tags introducing more semantic information from images, which further constrains LLMs and reduces hallucinations~(as shown in Fig.~\ref{fig:capsfusion_sample}).

\begin{table}[h]
\centering
\resizebox{\linewidth}{!}{
    \begin{tabular}{lccc}
        \toprule

        Method  & \shortstack{Text Generation\\ Model}  & \shortstack{Linear probe\\ Avg} & \shortstack{Zero-shot\\ Avg} \\
        \midrule
        RWKV-CLIP-B/32& CapsFusion  & 72.6 & 33.1  \\
        RWKV-CLIP-B/32& Ours  & \bf73.5 & \bf35.2 \\
	\bottomrule
    \end{tabular}
    }
\caption{Performance comparison using text generated by our proposed diverse description generation framework vs. CapsFusion.}
\label{tab:capsfusion_comparision}
\vspace{-3mm}
\end{table}

\begin{table}[h!]
\centering
\resizebox{\linewidth}{!}{
    \begin{tabular}{cccc|cc}
    \toprule
    $T_{r}$ & $T_{s}$ & $T_{g}$ & Dataset & \shortstack{Linear probe\\ Avg}  & \shortstack{Zero-shot\\ Avg} \\
    \midrule
    \Checkmark & \XSolidBrush &\XSolidBrush &YFCC15M& 71.3  & 38.7  \\
    \XSolidBrush & \Checkmark  & \XSolidBrush &YFCC15M& 72.4  & 23.1  \\
    \XSolidBrush & \XSolidBrush & \Checkmark &YFCC15M& 73.5  & 35.2  \\
    \Checkmark & \Checkmark & \XSolidBrush &YFCC15M& 73.0  & 43.0  \\
    \Checkmark & \XSolidBrush & \Checkmark &YFCC15M& 73.8  & 43.4  \\
    \Checkmark & \Checkmark & \Checkmark &YFCC15M& \bf74.1  & \bf44.4  \\
    \bottomrule
    \end{tabular}
 } 
\caption{Ablation experiment results using different types of text. $T_{r}$: raw text. $T_{s}$: synthetic caption. $T_{g}$: generated diverse description using our proposed framework.}
\label{tab:ablation-text}
\vspace{-5mm}
\end{table}

\begin{figure}[t]
\centering
{
\includegraphics[height=0.23\textwidth]{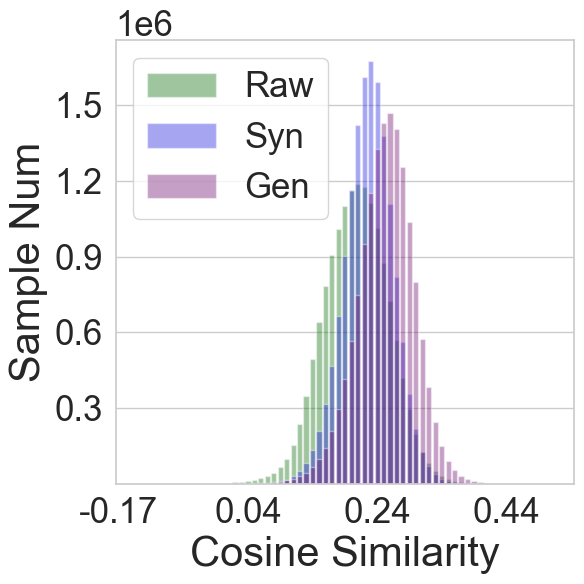}}
{
\includegraphics[height=0.23\textwidth]{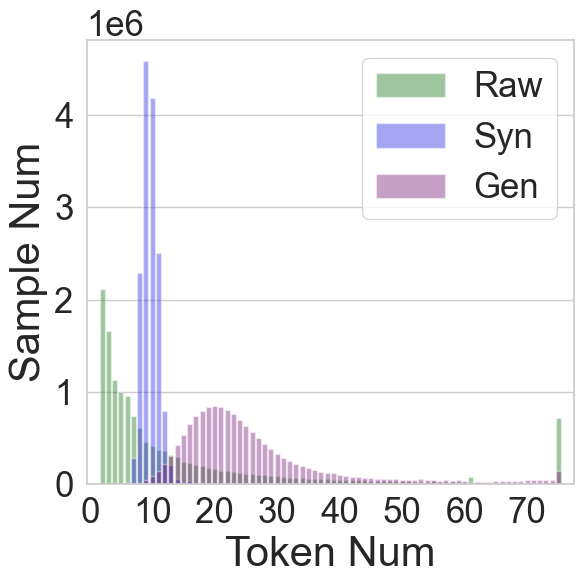}}
\caption{Statistical analysis of raw texts, synthetic captions, and generated diverse descriptions on the YFCC15M.}
\vspace{-5mm}
\label{similarity_token}
\end{figure}

\noindent{\bf Ablation on Different Types of Text.} We conduct ablation experiments on different categories of text, the average linear probe results on 10 datasets and the average zero-shot classification accuracy on 11 datasets are shown in Tab.~\ref{tab:ablation-text}. Synthetic captions and generated diverse descriptions yielded superior linear probe performance compared to raw texts. This improvement is attributed to the high incidence of mismatched image-text pairs in raw texts, which can adversely affect representation learning. As shown in Fig.~\ref{similarity_token}, our analysis of cosine similarity (computed by CLIP-L14) and token counts across different text types reveals that synthetic captions and generated diverse descriptions have higher average similarity and token counts than raw texts. Furthermore, despite these advantages, raw texts achieve superior zero-shot classification results, mainly due to the constraints imposed by the prompt template.

\subsection{Ablation on Model Architecture}
In Tab.~\ref{tab:ablation-architecture}, base on text augmentation, we perform an ablation study combining RWKV and Transformer architectures. Compared with Transformer$_{I}$ and Transformer$_{T}$, the integration of RWKV$_{I}$ and Transformer$_{T}$ achieves a 2.7\% improvement on linear probe but the zero-shot classification performance declines by 10.8\%. This reduction is primarily due to the poor compatibility between the RWKV and Transformer architectures. Conversely, the combination of RWKV$_{I}$ and RWKV$_{T}$ yields improvements of 3.2\% and 2.7\% in linear probe and zero-shot classification, respectively, indicating that RWKV outperforms Transformer in vision-language representation learning.

\begin{table}[h!]
\centering
\resizebox{\linewidth}{!}{
    \begin{tabular}{cc|cc|cc}
    \toprule
     
    \multicolumn{2}{c|}{Image} & \multicolumn{2}{c|}{Text} & Linear Probe & Zero-shot  \\
     RWKV$_{I}$  & Transformer$_{I}$ & RWKV$_{T}$ & Transformer$_{T}$  & Avg & Avg \\ 
    \midrule
    \XSolidBrush & \Checkmark & \XSolidBrush & \Checkmark & 70.9  & 41.7   \\
    \Checkmark & \XSolidBrush & \XSolidBrush & \Checkmark & 73.6  & 30.9   \\
    \XSolidBrush & \Checkmark & \Checkmark & \XSolidBrush & 71.0  & 41.1   \\
    \Checkmark & \XSolidBrush & \Checkmark & \XSolidBrush & \bf 74.1  & \bf 44.4  \\
    \bottomrule
    \end{tabular}
 } 
\caption{Ablation on model architecture.}
\label{tab:ablation-architecture}
\vspace{-5mm}
\end{table}

\begin{figure}[h]
\centering
    {
    {\includegraphics[width=0.23\textwidth]{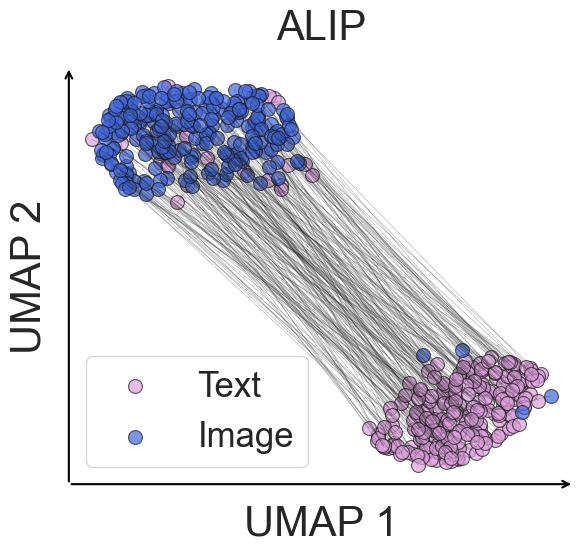}}}
    {
    {\includegraphics[width=0.23\textwidth]{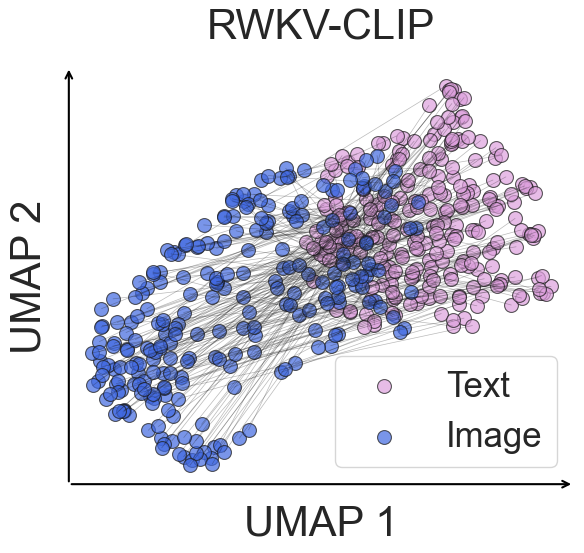}}}
\caption{Visualization of modality gaps.}
\label{fig:modality_gap}
\vspace{-5mm}
\end{figure}

\noindent{\bf Analysis of Feature Embedding.} To understand what makes RWKV-CLIP effective, we randomly select 250 image-text pairs from YFCC15M and visualize the modality gaps of ALIP and RWKV-CLIP. Specifically, each image and its corresponding text are encoded into embedding space and reduced to two dimensions using UMAP~\cite{mcinnes2018umap}. As shown in Fig.~\ref{fig:modality_gap}, we found that the representations learned by RWKV-CLIP exhibit clearer discriminability within the same modality. Additionally, compared to ALIP, RWKV-CLIP demonstrates closer distances in the image-text modality space, indicating superior cross-modal alignment performance. 

%% file: section/conclusion.tex
\section{Conclusion}
In this paper, we further explore CLIP from the perspectives of data and model architecture. We introduce a diverse description generation framework that can leverage Large Language Models~(LLMs) to combine and refine information from web-based image-text pairs, synthetic captions, and detection tags. Besides, we propose RWKV-CLIP, the first RWKV-driven vision-language representation learning model that combines the effective parallel training of transformers with the efficient inference of RNNs. Our method demonstrates superior performance across various model scales and pre-training datasets on different downstream tasks. We hope that our work provides insights into vision-language representation learning models.

%% file: appendix.tex
\appendix

\section{Detail Experimental Settings}
\subsection{Model Architectures}
We meticulously regulate the parameters and FLOPs of RWKV-CLIP to ensure the fairness of the experimental comparison. The detailed parameters and FLOPs of RWKV-CLIP-B/32 and RWKV-CLIP-B/16 are shown in Tab.~\ref{tab:params}. The detail settings of RWKV-CLIP-B/32 and RWKV-CLIP-B/16 are shown in Tab.~\ref{tab:model_settings}. 

\begin{table}[h]
\centering
\resizebox{\linewidth}{!}{
    \begin{tabular}{l|cc|cc|cc}
        \toprule
        \multirow{2}{*}{Method} & \multicolumn{2}{c|}{Image} & \multicolumn{2}{c|}{Text} & \multicolumn{2}{c}{Total} \\
        & \small Params(M)  & \small FLOPs(G) & \small Params(M)  & \small FLOPs(G) & \small Params(M)  & \small FLOPs(G)  \\
        \midrule
        \color{dt}\small CLIP-ViT-B/32  & \color{dt} 87.85 & \color{dt} 8.73  & \color{dt} 63.44 & \color{dt} 5.82 & \color{dt} 151.29 & \color{dt} 14.55  \\
        \small RWKV-CLIP-B/32  &  84.21 & 7.91  &  65.35 & 4.93 &  149.56 & 12.84 \\
	\midrule
        \color{dt}\small CLIP-ViT-B/16  & \color{dt} 86.19 & \color{dt} 33.72 &  \color{dt} 63.44 & \color{dt} 5.82 & \color{dt} 149.63 & \color{dt} 39.54 \\
        \small RWKV-CLIP-B/16  & 82.83 & 31.05 &  65.35 & 4.93 & 148.18 & 35.98 \\
	\bottomrule
    \end{tabular}
    }
\caption{Parameters and FLOPs comparison between CLIP and RWKV-CLIP.}
\vspace{-5mm}
\label{tab:params}
\end{table}

\subsection{Detail Instruction Prompt}
The prompt used to input ChatGPT is present in the following:
\\
\emph{\color{black!65}"Please merge the information from the given raw text and the synthetic caption with the help of the highly relevant detection tags. The raw caption offers detailed real-world information, yet it suffers from flaws in sentence structure and grammar. The synthetic caption exhibits impeccable sentence structure but often lacks in-depth real-world details and may contain false information. The highly relevant detection tags are provided to enrich the semantic information of the raw caption, while some are redundant and noisy. You are a great information integration and summary expert, you are also good at enriching semantic information. Ensure a well-structured sentence while retaining the detailed real-world information provided in the raw caption. Avoid simply concatenating the sentences and avoid adding external information to describe. Correctness and simplify sentences finally. Raw caption:<raw caption>, synthetic caption:<synthetic caption>, and highly relevant detection tags:<detection tags>".}

\subsection{Experimental Settings}
We present the settings used in the training RWKV-CLIP in Tab.~\ref{tab:hyperparams}.

\begin{table}[h]
    \centering
    \resizebox{0.8\linewidth}{!}{
     \begin{tabular}{l|c}
        \toprule Hyperparameter & Value \\
        \midrule
        Initial temperature & $0.07$ \\
        Adam $\beta_{1}$ & $0.9$ \\
        Adam $\beta_{2}$ & $0.98$ \\
        Adam $\epsilon$  & $10^{-6}$ \\
        Weight decay & $0.2$ \\
        Batch size & 4096 \\
        Learning rate & 0.001 \\
        Learning rate scheduler & OneCycleLR \\
        Pct start & 0.1 \\
        Training epochs & 32  \\
        GPU & $8 \times $A100 \\
        \bottomrule
    \end{tabular}
    }
\centering
\caption{Hyperparameters used for RWKV-CLIP pre-training.}
\label{tab:hyperparams}
\vspace{-5mm}
\end{table}

\subsection{Prompts for Zero-shot Classification}
In this work, we evaluate the zero-shot performance of RWKV-CLIP on 11 downstream datasets. All the prompts for the 11 downstream datasets are presented in Tab.~\ref{tab:prompt}.

\begin{table*}[h!]
\centering
\resizebox{0.9\linewidth}{!}{
\begin{tabular}{c|cc|cccc|cccc} \toprule
       & Embedding & Input      & \multicolumn{4}{c|}{Image Encoder} & \multicolumn{4}{c}{Text Encoder} \\
    Model  & dimension & resolution & layers & hidden rate & heads & Init & layers  & hidden rate & heads & Init \\ \midrule
    RWKV-CLIP-B/32  & 640 & 224 & 12 & 5 & 8 &\checkmark & 6  & 3.5 & 10 &\checkmark \\
    RWKV-CLIP-B/16  & 640 & 224 & 12 & 5 & 8 &\checkmark & 6  & 3.5 & 10 &\checkmark \\
    \bottomrule
\end{tabular}
}
\caption{The detail architecture parameters for our proposed RWKV-CLIP.}
\label{tab:model_settings}
\vspace{-3mm}
\end{table*}

\begin{table*}[h!]
	\centering
	\resizebox{\linewidth}{!}{
		\begin{tabular}{llcccccccccccccccccccccccccccc}
			\toprule
			 & Method              & \shortstack{Pre-train\\ data} & \rotatebox[origin=lb]{90}{\smash{Food101}} & \rotatebox[origin=lb]{90}{\smash{CIFAR10}} & \rotatebox[origin=lb]{90}{\smash{CIFAR100}} & \rotatebox[origin=lb]{90}{\smash{Birdsnap}} & \rotatebox[origin=lb]{90}{\smash{SUN397}} & \rotatebox[origin=lb]{90}{\smash{Cars}} & \rotatebox[origin=lb]{90}{\smash{Aircraft}} & \rotatebox[origin=lb]{90}{\smash{VOC2007}} & \rotatebox[origin=lb]{90}{\smash{DTD}} & \rotatebox[origin=lb]{90}{\smash{Pets}} & \rotatebox[origin=lb]{90}{\smash{Caltech101}} & \rotatebox[origin=lb]{90}{\smash{Flowers}} & \rotatebox[origin=lb]{90}{\smash{MNIST}} & \rotatebox[origin=lb]{90}{\smash{STL10}} & \rotatebox[origin=lb]{90}{\smash{EuroSAT}} & \rotatebox[origin=lb]{90}{\smash{RESISC45}} & \rotatebox[origin=lb]{90}{\smash{GTSRB}} & \rotatebox[origin=lb]{90}{\smash{KITTI}} & \rotatebox[origin=lb]{90}{\smash{Country211}} & \rotatebox[origin=lb]{90}{\smash{PCAM}} & \rotatebox[origin=lb]{90}{\smash{UCF101}} & \rotatebox[origin=lb]{90}{\smash{Kinetics700}} & \rotatebox[origin=lb]{90}{\smash{CLEVR}} & \rotatebox[origin=lb]{90}{\smash{Memes}} & \rotatebox[origin=lb]{90}{\smash{SST2}} &
             \rotatebox[origin=lb]{90}{\smash{ImageNet}} &
             \rotatebox[origin=lb]{90}{\smash{Average}} \\

              \midrule  
              & ALIP-ViT-B/32     & LAION10M & 71.5 	& 92.2 	& 76.1 	& 36.3 	& 67.3 	& 70.1 	& 41.8 	& 85.3 	& 71.3 	& 74.3 	& 86.9 	& \bf 90.7 	& 98.0 	& 94.6 	& 95.4 	& 84.3 	& 84.1 	& 70.0 	& 12.9 	& 83.4 	& 75.9 	& 46.4 	& 51.0 	& 54.8 	& 56.5 	& 59.6 & 70.4   \\   
              
              & RWKV-CLIP-B/32     & LAION10M & \bf 72.7 & \bf 94.7 & \bf 81.4 & \bf 42.3 & \bf 68.3 & \bf 70.3 & \bf 47.9 & \bf 86.5 & \bf 73.6 & \bf 76.6 & \bf 90.0 & 89.4 & \bf 99.0 & \bf 94.6 & \bf 97.0 & \bf 85.6 & \bf 87.0 & \bf 74.9 & \bf 13.8 & \bf 85.1 & \bf 80.8 & \bf 49.3 & \bf 60.6 & \bf 55.4 & \bf 58.3 & \bf 63.7 & \bf 73.0   \\  
              \midrule  
                   
              & ALIP-ViT-B/16     & LAION10M & 77.2 	& 93.3 	& 77.0 	& 45.1 	& 69.4 	& \bf 77.3 	& 48.6 	& 87.7 	& 74.5 	& 79.0 	& 88.1 	& \bf 93.0 	& 98.3 	& 96.3 	& 96.3 	& 86.4 	& 83.7 	& 72.2 	& 14.2 	& 85.2 	& 80.1 	& 50.1 	& 55.4 	& 55.7 	& 57.3 	& 64.8  & 73.3  \\    
              & RWKV-CLIP-B/16    & LAION10M & \bf 78.9 & \bf 95.1 & \bf 81.8 & \bf 50.3 & \bf 72.0 & 76.8 & \bf 50.3 & \bf 89.4 & \bf 75.4 & \bf 79.7 & \bf 91.9 & 91.7 & \bf 99.0 & \bf 96.4 & \bf 96.9 & \bf 87.8 & \bf 87.4 & \bf 75.7 & \bf 15.2 & \bf 85.5 & \bf 83.9 & \bf 53.0 & \bf 61.8 & \bf 55.9 & \bf 60.0 & \bf 68.4 & \bf 75.4 \\ 
              
              \midrule           

              & ALIP-ViT-B/32  & LAION30M & 76.6 	& 94.0 	& 79.3 	& 44.2 	& 70.6 	& 77.7 	& 48.4 	& 87.6 	& 74.4 	& \bf 80.4 	& 90.0 	& \bf 93.8 	& 98.3 	& 96.3 	& 96.0 	& 86.7 	& 84.7 	& 72.3 	& 15.0 	& 85.0 	& 81.0 	& 50.6 	& 55.6 	& 56.1 	& \bf 59.8 	& 65.0  & 73.8  \\ 

              & RWKV-CLIP-B/32 & LAION30M & \bf 76.6 & \bf 95.6 & \bf 82.8 & \bf 46.0 & \bf 71.0 & \bf 77.9 & \bf 50.0 & \bf 88.2 & \bf 74.5 & 78.9 & \bf 91.6 & 92.1 & \bf 99.0 & \bf 96.5 & \bf 97.1 & \bf 86.9 & \bf 87.6 & \bf 78.9 & \bf 15.2 & \bf 85.6 & \bf 83.4 & \bf 51.8 & \bf 61.6 & \bf 58.9 & 58.9 & \bf 67.2 & \bf 75.2  \\  
            \bottomrule
		\end{tabular}
    }
  \caption{Top-1 accuracy(\%) of linear probe on 26 image classification datasets. }
  \vspace{-5mm}
  \label{tab:linear_probe}
\end{table*}

\section{Detail Linear Probe on LAION}

\begin{table}[t]
\centering
\resizebox{\linewidth}{!}{
\begin{tabular}{lcccr}
\toprule
\multicolumn{1}{l}{Dataset} & \multicolumn{1}{c}{Classes} & \multicolumn{1}{c}{Train size} & \multicolumn{1}{c}{Test size} & \multicolumn{1}{c}{Evaluation metric} \\
\midrule
Food101                        & 102                             & 75,750                             & 25,250                            & accuracy                                  \\
CIFAR10                        & 10                              & 50,000                             & 10,000                            & accuracy                                  \\
CIFAR100                       & 100                             & 50,000                             & 10,000                            & accuracy                                  \\
Birdsnap                        & 500                             & 42,138                             & 2,149                             & accuracy                                  \\
SUN397                          & 397                             & 19,850                             & 19,850                            & accuracy                                  \\
Cars                   & 196                             & 8,144                              & 8,041                             & accuracy                                  \\
Aircraft                   & 100                             & 6,667                              & 3,333                             & mean per class                            \\
VOC2007                   & 20                             & 5011                             & 4952                             & 11-point mAP                            \\
DTD           & 47                              & 3,760                              & 1,880                             & accuracy                                  \\
Pets                & 37                              & 3,680                              & 3,669                             & mean per class                            \\
Caltech101                     & 101                             & 3,000                              & 5,677                             & mean-per-class                            \\
Flowers                  & 102                             & 2,040                              & 6,149                             & mean per class                            \\
MNIST                  & 10                             & 60,000                              & 10,000                            & accuracy                            \\
STL10                        & 10                             & 5,000                             & 8,000                            & accuracy                                  \\
EuroSAT                         & 10                              & 10,000                             & 5,000                             & accuracy                                  \\
RESISC45            & 45                              & 3,150                              & 25,200                             & accuracy                                  \\
GTSRB            & 43                 &26,640                 &12,630                                                   & accuracy                                  \\
KITTI            & 4                              & 6770                             & 711                             & accuracy                                  \\
Country211            & 211                              & 42,200                             & 21,100                             & accuracy                                  \\
PCAM            & 2                              & 294,912                             & 32,768                             & accuracy                                  \\
UCF101            & 101                              & 9,537                             & 1,794                            & accuracy                                  \\
Kinetics700            & 700                              & 530,779                             & 33,944                            & mean(top1,top5)                                  \\
CLEVR            & 8                              & 2,000                             & 500                            & accuracy                                  \\
Memes            & 2                              & 8,500                             & 500                            & ROC AUC                                  \\
SST2            & 2                              & 7,792                             & 1,821                            & accuracy                                  \\
ImageNet                        & 1000                            & 1,281,167                          & 50,000                            & accuracy                                  \\
\bottomrule
\end{tabular}}
\caption{List of linear probe datasets with the data distribution and evaluation metrics.}
\vspace{-3mm}
\label{linearprobe_datasets}
\end{table}

\subsection{Downstream Datasets}
To comprehensively demonstrate the performance of RWKV-CLIP, we compared the linear probe results of RWKV-CLIP and ALIP across 26 datasets. These datasets
include Food101~\cite{bossard2014food}, CIFAR10~\cite{krizhevsky2009learning}, CIFAR100~\cite{krizhevsky2009learning}, 
Birdsnap~\cite{berg2014birdsnap},
SUN397~\cite{xiao2010sun},
Stanford Cars~\cite{KrauseStarkDengFei-Fei_3DRR2013},
FGVC Aircraft~\cite{maji2013fine},
VOC2007~\cite{everingham2007pascal},
DTD~\cite{cimpoi2014describing},
Pets~\cite{parkhi2012cats}, 
Caltech101~\cite{fei2004learning},
Flowers102~\cite{nilsback2008automated},
MNIST~\cite{lecun1998gradient},
SLT10~\cite{coates2011analysis},
EuroSAT~\cite{helber2019eurosat},
RESISC45~\cite{cheng2017remote},
GTSRB~\cite{stallkamp2012man},
KITTI~\cite{geiger2012we},
Country211~\cite{radford2021learning},
PCAM~\cite{veeling2018rotation},
UCF101~\cite{soomro2012ucf101},
Kinetics700~\cite{carreira2019short},
CLEVR~\cite{johnson2017clevr},
Hateful Memes~\cite{kiela2020hateful},
SST2~\cite{radford2021learning},
ImageNet~\cite{deng2009imagenet}. Details on each dataset and the corresponding evaluation metrics are provided in Tab.~\ref{linearprobe_datasets}.

\subsection{Detail Linear Probe Results}
Following ALIP, we conduct experiments on randomly selected subsets of 10M and 30M from the LAION400M dataset. For a comprehensive comparison, we report the linear probe performance on 26 downstream datasets. The complete experimental results are shown in Tab.\ref{tab:linear_probe}. RWKV-CLIP-B/32 outperforms ALIP-ViT-B/32 2.6\% and 1.4\% when training on LAION10M and LAION30M, respectively. Additionally, RWKV-CLIP-B/16 also surpasses ALIP-ViT-B/16 by 2.1\% on average across the 26 datasets. These experimental results indicate that RWKV-CLIP demonstrates both robustness and extensibility.

\section{More Visualize and Analysis}

\subsection{Class Activation Map}
\begin{figure}[t]
\centering
  \includegraphics[width=\linewidth]{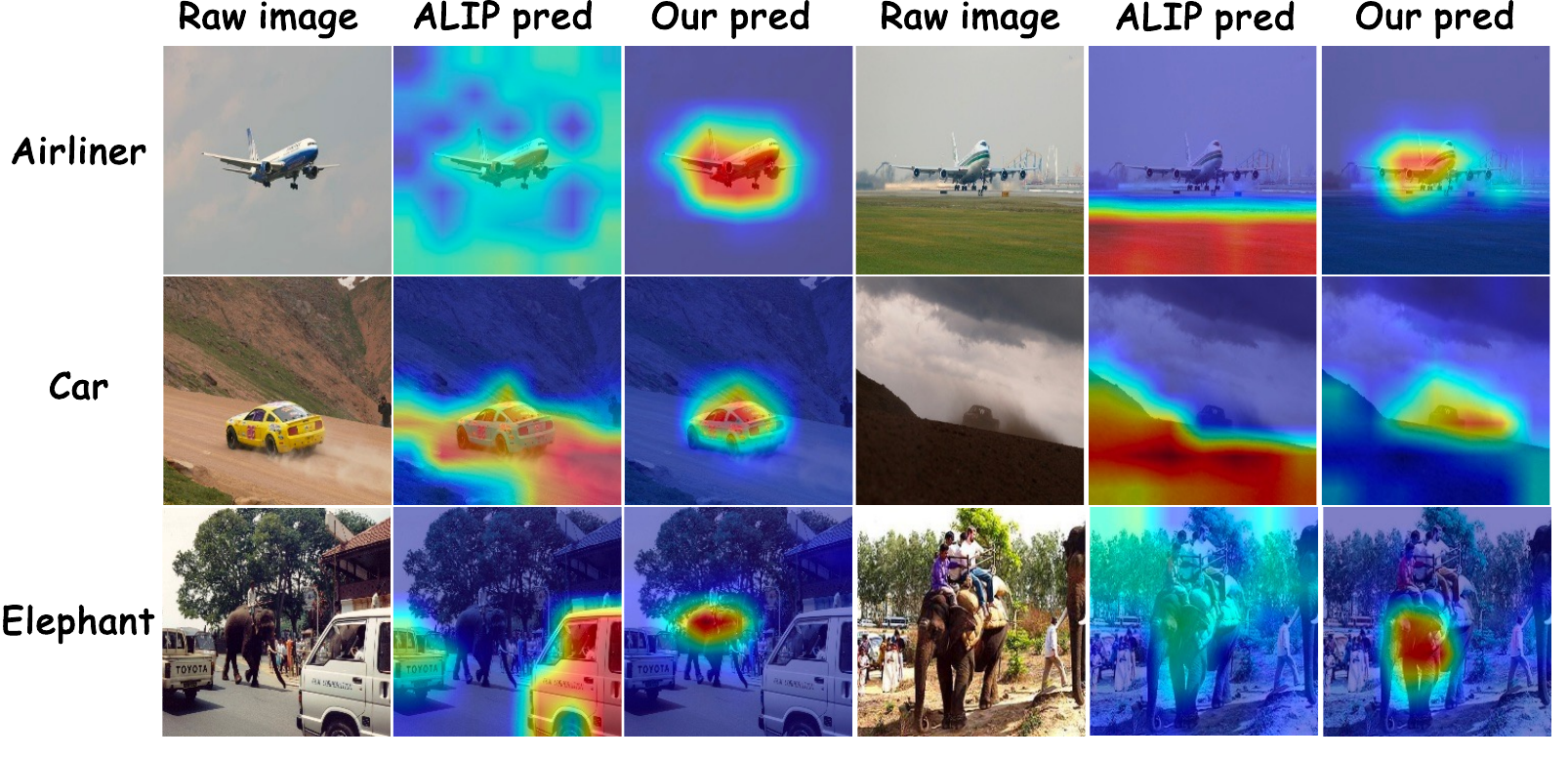}
  \vspace{-3mm}
  \caption{Class activation maps for ALIP and RWKV-CLIP on different classes from ImageNet.}
  \label{fig:activation_map}
\vspace{-3mm}
\end{figure}

As shown in Fig.~\ref{fig:activation_map}, we visualize the class activation maps of ALIP and RWKV-CLIP on different classes from ImageNet. RWKV-CLIP performs superior in aligning the image patches and textual tokens. For example, RWKV-CLIP captures corresponding text semantic entities in images more accurately.

\subsection{Cross Modal Alignment Analysis}
\begin{figure}[h]
\centering
    {
    {\includegraphics[width=0.23\textwidth]{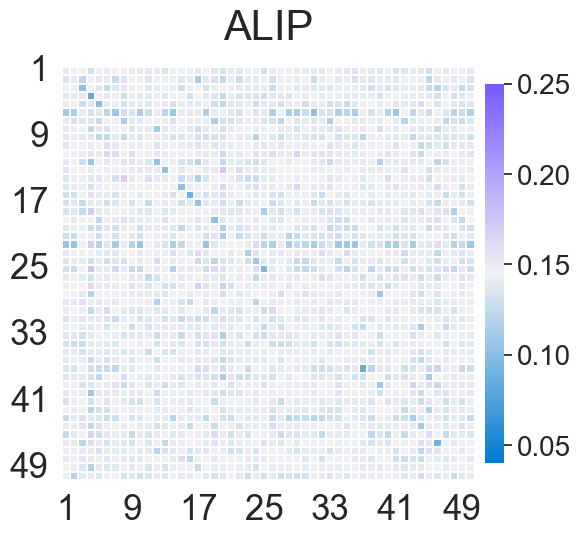}}}
    {
    {\includegraphics[width=0.23\textwidth]{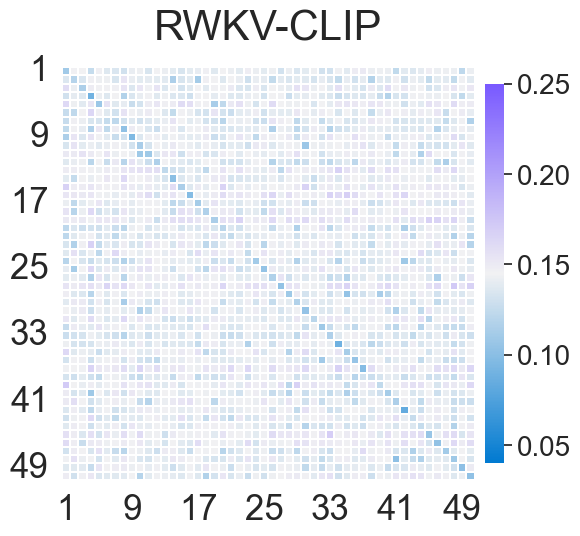}}}
\caption{Visualization of modality gaps.}
\label{fig:alignment_compare}
\vspace{-3mm}
\end{figure}

To evaluate the performance of the cross-modal alignment of RWKV-CLIP, we random select 50 samples from YFCC15M and visualize the cross-modal cosine similarity matrix in Fig.~\ref{fig:alignment_compare}. We observe that the diagonal of the RWKV-CLIP matrix is significantly clearer compared to ALIP, indicating that the representations learned by RWKV-CLIP exhibit greater distinctiveness and improved cross-modal alignment capability.

\subsection{Case Study}
\begin{figure*}[h!]
\centering
  \includegraphics[width=0.9\linewidth]{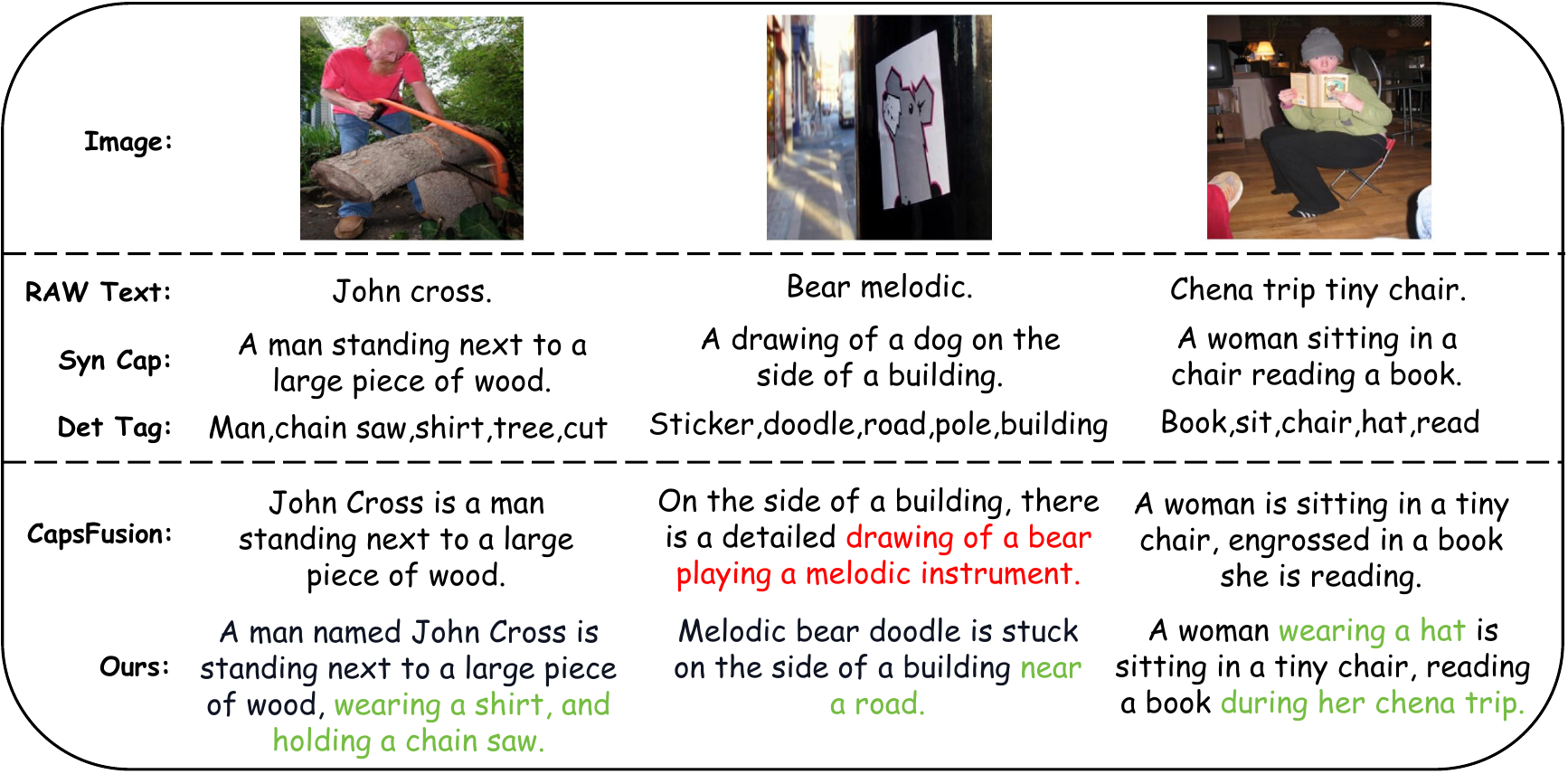}
  \caption{Comparison of generated text using our proposed diverse description generation framework vs. CapsFusion. Hallucinations are highlighted in \textcolor[HTML]{FF1E1E}{red}, and additional semantic information is highlighted in \textcolor[HTML]{6FBF50}{green}.}
  \label{fig:case}
\end{figure*}

In Fig.~\ref{fig:case}, we visualize additional generated text using CapsFusion and our proposed framework. The introduction of detection tags enhances semantic information from images, thereby constraining LLMs and significantly reducing hallucinations.

\begin{table*}[h!]
\centering
\resizebox{0.85\linewidth}{!}{
\begin{tabular}{llll}
\toprule
\multicolumn{4}{l}{\bf CIFAR 10 \& CIFAR 100} \\
a photo of a \{label\}. &
a blurry photo of a \{label\}. &
a black and white photo of a \{label\}. &
a low contrast photo of a \{label\}. \\
a high contrast photo of a \{label\}. &
a bad photo of a \{label\}. &
a good photo of a \{label\}. &
a photo of a small \{label\}. \\
a photo of a big \{label\}.&
a photo of the \{label\}.&
a blurry photo of the \{label\}.&
a black and white photo of the \{label\}. \\
a low contrast photo of the \{label\}.&
a high contrast photo of the \{label\}.&
a bad photo of the \{label\}.&
a good photo of the \{label\}. \\
a photo of the small \{label\}.&
a photo of the big \{label\}.& & \\
\midrule
\multicolumn{4}{l}{\bf Food101} \\
a photo of \{label\}, a type of food. & & \\
\midrule
\multicolumn{4}{l}{\bf Caltech101} \\
a photo of a \{label\}. &
a painting of a \{label\}. &
a plastic \{label\}. &
a sculpture of a \{label\}. \\
a sketch of a \{label\}. &
a tattoo of a \{label\}. &
a toy \{label\}. &
a rendition of a \{label\}. \\
a embroidered \{label\}. &
a cartoon \{label\}. &
a \{label\} in a video game. &
a plushie \{label\}. \\
an origami \{label\}. &
art of a \{label\}. &
graffiti of a \{label\}. &
a drawing of a \{label\}. \\
a doodle of a \{label\}. &
a photo of the \{label\}. &
a painting of the \{label\}.&
the plastic \{label\}. \\
a sculpture of the \{label\}.&
a sketch of the \{label\}.&
a tattoo of the \{label\}.&
the toy \{label\}. \\
a rendition of the \{label\}.&
the embroidered \{label\}.&
the cartoon \{label\}.&
the \{label\} in a video game. \\
the plushie \{label\}.&
the origami \{label\}.&
art of the \{label\}.&
graffiti of the \{label\}. \\
a drawing of the \{label\}.&
a doodle of the \{label\}.& & \\
\midrule
\multicolumn{4}{l}{\bf Stanford Cars} \\
a photo of a \{label\}.&
a photo of the \{label\}.&
a photo of my \{label\}.&
i love my \{label\}! \\
a photo of my dirty \{label\}.&
a photo of my clean \{label\}.&
a photo of my new \{label\}.&
a photo of my old \{label\}. \\
\midrule
\multicolumn{4}{l}{\bf DTD} \\
a photo of a \{label\} texture.&
a photo of a \{label\} pattern.&
a photo of a \{label\} thing.&
a photo of a \{label\} object. \\
a photo of the \{label\} texture. &
a photo of the \{label\} pattern. &
a photo of the \{label\} thing. &
a photo of the \{label\} object. \\
\midrule
\multicolumn{4}{l}{\bf FGVC Aircraft} \\
a photo of a \{label\}, a type of aircraft.&
a photo of the \{label\}, a type of aircraft.& & \\
\midrule
\multicolumn{4}{l}{\bf Flowers102} \\
a photo of a \{label\}, a type of flower. &&& \\
\midrule
\multicolumn{4}{l}{\bf Pets } \\
a photo of a \{label\}, a type of pet.&&& \\
\midrule
\multicolumn{4}{l}{\bf  SUN39} \\
a photo of a \{label\}.&
a photo of the \{label\}.&& \\
\midrule
\multicolumn{4}{l}{\bf  ImageNet} \\
a bad photo of a \{label\}. & 
a photo of many \{label\}. &
a sculpture of a \{label\}. &
a photo of the hard to see \{label\}. \\
a low resolution photo of the \{label\}. & 
a rendering of a \{label\}. &
graffiti of a \{label\}. &
a bad photo of the \{label\}.  \\
a cropped photo of the \{label\}. &
a tattoo of a \{label\}. & 
the embroidered \{label\}. &
a photo of a hard to see \{label\}.  \\
a bright photo of a \{label\}.&
a photo of a clean \{label\}.&
a photo of a dirty \{label\}.&
a dark photo of the \{label\}. \\
a drawing of a \{label\}.&
a photo of my \{label\}.&
the plastic \{label\}.&
a photo of the cool \{label\}. \\
a close-up photo of a \{label\}.&
a black and white photo of the \{label\}.&
a painting of the \{label\}.&
a painting of a \{label\}. \\
a pixelated photo of the \{label\}.& 
a sculpture of the \{label\}.&
a bright photo of the \{label\}.&
a cropped photo of a \{label\}. \\
a plastic \{label\}.&
a photo of the dirty \{label\}.& 
a jpeg corrupted photo of a \{label\}.&
a blurry photo of the \{label\}. \\
a photo of the \{label\}.&
a good photo of the \{label\}.&
a rendering of the \{label\}.&
a \{label\} in a video game. \\
a photo of one \{label\}.&
a doodle of a \{label\}.&
a close-up photo of the \{label\}.&
a photo of a \{label\}. \\
the origami \{label\}.&
the \{label\} in a video game.&
a sketch of a \{label\}.&
a doodle of the \{label\}. \\
an origami \{label\}.&
a low resolution photo of a \{label\}.&
the toy \{label\}.&
a rendition of the \{label\}. \\
a photo of the clean \{label\}.& 
a photo of a large \{label\}.& 
a rendition of a \{label\}.&
a photo of a nice \{label\}. \\
a photo of a weird \{label\}.& 
a blurry photo of a \{label\}.&
a cartoon \{label\}.&
art of a \{label\}. \\
a sketch of the \{label\}.& 
a embroidered \{label\}.&
a pixelated photo of a \{label\}.&
itap of the \{label\}. \\
a jpeg corrupted photo of the \{label\}.& 
a good photo of a \{label\}.&
a plushie \{label\}.&
a photo of the nice \{label\}. \\
a photo of the small \{label\}.& 
a photo of the weird \{label\}.&
the cartoon \{label\}.&
art of the \{label\}. \\
a drawing of the \{label\}.& 
a photo of the large \{label\}.& 
a black and white photo of a \{label\}.&
the plushie \{label\}. \\
a dark photo of a \{label\}.& 
itap of a \{label\}.& 
graffiti of the \{label\}.& 
a toy \{label\}. \\
itap of my \{label\}.& 
a photo of a cool \{label\}.&
a photo of a small \{label\}.& 
a tattoo of the \{label\}. \\
\bottomrule
\end{tabular}}
\caption{Full list of prompts to evaluate the performance of zero-shot classification on 11 visual recognition datasets.}
\label{tab:prompt}
\end{table*}